\documentclass{article}
\usepackage[accepted]{icml2018}
\usepackage{xcolor}
\usepackage{soul}
\usepackage{times}  
\usepackage{helvet}  
\usepackage{courier}  
\usepackage{url}  
\usepackage{graphicx}  

\usepackage[utf8]{inputenc}
\usepackage[T1]{fontenc}
\usepackage{amsmath}
\usepackage{bbold}
\usepackage{comment}
\usepackage{times}
\usepackage{booktabs}
\usepackage{footnote}
\usepackage{verbatim}
\usepackage[colorlinks=true]{hyperref}
\hypersetup{
  colorlinks = true,
  citecolor = blue,
  pdftitle={Asymmetric Variational Autoencoders},
  pdfauthor={Guoqing Zheng, Yiming Yang, Jaime Carbonell}
}
\usepackage{caption, subcaption}
\usepackage{microtype}
\usepackage{tablefootnote}

\usepackage{tikz}
\usetikzlibrary{positioning}

\newcommand{\vect}[1]{\mathbf{#1}}

\newtheorem{theorem}{Theorem}

\newtheorem{proposition}[theorem]{Proposition}

\title{Asymmetric Variational Autoencoders}

\icmltitlerunning{Asymmetric Variational Autoencoders}

\begin{document}

\twocolumn[
\icmltitle{Asymmetric Variational Autoencoders}



 
\begin{icmlauthorlist}
  \icmlauthor{Guoqing Zheng}{cmu}
  \icmlauthor{Yiming Yang}{cmu}
  \icmlauthor{Jaime Carbonell}{cmu}
\end{icmlauthorlist}

\icmlaffiliation{cmu}{School of Computer Science, Carnegie Mellon University, Pittsburgh PA, USA}
\icmlcorrespondingauthor{Guoqing Zheng}{gzheng@cs.cmu.edu}

\icmlkeywords{Machine Learning, ICML}

\vskip 0.3in
]



\printAffiliationsAndNotice{}  

\begin{abstract}

  Variational inference for latent variable models is prevalent in
  various machine learning problems, typically solved by maximizing
  the Evidence Lower Bound (ELBO) of the true data likelihood with
  respect to a variational distribution. However, freely enriching the
  family of variational distribution is challenging since the ELBO
  requires variational likelihood evaluations of the latent
  variables. In this paper, we propose a novel framework to enrich the
  variational family by incorporating auxiliary variables to the
  variational family. The resulting inference network doesn't require
  density evaluations for the auxiliary variables and thus complex
  implicit densities over the auxiliary variables can be constructed
  by neural networks. It can be shown that the actual variational
  posterior of the proposed approach is essentially modeling a rich
  probabilistic mixture of simple variational posterior indexed by
  auxiliary variables, thus a flexible inference model can be
  built. Empirical evaluations on several density estimation tasks
  demonstrates the effectiveness of the proposed method.

\end{abstract}

\section{Introduction}

Estimating posterior distributions is the primary focus of Bayesian
inference, where we are interested in how our belief over the
variables in our model would change after observing a set of data.
Predictions can also be benefited from Bayesian inference as every
prediction will be equipped with a confidence interval representing
how sure the prediction is.  Compared to the maximum a posteriori
(MAP) estimator of the model parameters, which is a point estimator,
the posterior distribution provides richer information about model
parameters and hence more justified prediction.

Among various inference algorithms for posterior estimation,
variational inference (VI)~\cite{blei2017variational} and Markov Chain
Monte Carlo (MCMC)~\cite{geyer1992practical} are the most wisely used
ones. It is well known that MCMC suffers from slow mixing time though
asymptotically the chained samples will approach the true
posterior. Furthermore, for latent variable models
(LVMs)~\cite{wainwright2008graphical} where each sampled data point is
associated with a latent variable, the number of simulated Markov
Chains increases with the number of data points, making the
computation too costly. VI, on the other hand, facilitates faster
inference because it optimizes an explicit objective function and its
convergence can be measured and controlled. Hence, VI has been widely
used in many Bayesian models, such as the mean-field approach for the
Latent Dirichlet Allocation \cite{DBLP:journals/jmlr/BleiNJ03},
etc. To enrich the family of distributions over the latent variables,
neural network based variational inference methods have also been
proposed, such as Variational Autoencoder
(VAE)~\cite{DBLP:journals/corr/KingmaW13}, Importance Weighted
Autoencoder (IWAE)~\cite{DBLP:journals/corr/BurdaGS15} and
others~\cite{DBLP:conf/icml/RezendeM15,DBLP:conf/icml/MnihG14,DBLP:conf/nips/KingmaSJCCSW16}. These
methods outperform the traditional mean-field based inference
algorithms due to their flexible distribution families and
easy-to-scale algorithms, therefore becoming the state of the art for
variational inference.

The aforementioned VI methods are essentially maximizing the evidence
lower bound (ELBO), i.e., the lower bound of the true marginal data
likelihood, defined as
\begin{align}
  \log p_\theta(x)&\geq \mathbb{E}_{z\sim q_\phi(z|x)}\log\frac{p(z,x)}{q(z|x)} \label{eq:2}
\end{align}
where $x, z$ are data point and its latent code, $p$ and $q$ denote
the generative model and the variational model, respectively. The
equality holds if and only if $q_\phi(z|x)=p_\theta(z|x)$ and
otherwise a gap always exists. The more flexible the variational
family $q(z|x)$ is, the more likely it will match the true posterior
$p(z|x)$. However, arbitrarily enriching the variational model family
$q$ is non-trivial, since optimizing Eq. \ref{eq:2} always requires
evaluations of $q(z|x)$. Most of existing methods either make over
simplified assumptions about the variational model, such as simple
Gaussian posterior in VAE~\cite{DBLP:journals/corr/KingmaW13}, or
resort to implicit variational models without explicitly modeling
$q(z|x)$~\cite{dumoulin2016adversarially}.

In this paper we propose to enrich the variational distribution
family, by incorporating auxiliary variables to the variational
model. \textit{Most importantly, density evaluations are not required
  for the auxiliary variables and thus complex implicit density over
  the auxiliary variables can be easily constructed, which in turn
  results in a flexible variational posterior over the latent
  variables}. We argue that the resulting inference network is
essentially modeling a complex probabilistic mixture of different
variational posteriors indexed by the auxiliary variable, and thus a
much richer and flexible family of variational posterior distribution
is achieved. We conduct empirical evaluations on several density
estimation tasks, which validate the effectiveness of the proposed
method.

The rest of the paper is organized as follows: We briefly review two
existing approaches for inference network modeling in Section
\ref{sec:prelim}, and present our proposed framework in the Section
\ref{sec:model}. We then point out the connections of the proposed
framework to related methods in Section \ref{sec:related}. Empirical
evaluations and analysis are carried out in Section \ref{sec:exp}, and
lastly we conclude this paper in the Section \ref{sec:conclusion}.

\section{Preliminaries}
\label{sec:prelim}

In this section, we briefly review several existing methods that aim
to address variational inference with stochastic neural networks.

\subsection{Variational Autoencoder (VAE)}

Given a generative model $p_\theta(x,z)=p_\theta(z)p_\theta(x|z)$
defined over data $x$ and latent variable $z$, indexed by parameter
$\theta$, variational inference aims to approximate the intractable
posterior $p(z|x)$ with $q_\phi(z|x)$, indexed by parameter $\phi$,
such that the ELBO is maximized
\begin{align}
  \mathcal{L}_{\textsc{VAE}}(x)\equiv\mathbb{E}_q\log p(x,z)-\mathbb{E}_q \log q(z|x)\leq \log p(x)
\end{align}
Parameters of both generative distribution $p$ and variational
distribution $q$ are learned by maximizing the ELBO with stochastic
gradient methods.\footnote{We drop the dependencies of $p$ and $q$ on
  parameters $\theta$ and $\phi$ to prevent clutter.}
Specifically, VAE~\cite{DBLP:journals/corr/KingmaW13} assumes both the
conditional distribution of data given the latent codes of the
generative model and the variational posterior distribution are
Gaussians, whose means and diagonal covariances are parameterized by
two neural networks, termed as generative network and inference
network, respectively. Model learning is possible due to the
re-parameterization trick~\cite{DBLP:journals/corr/KingmaW13} which
makes back propagation through the stochastic variables possible.

\subsection{Importance Weighted Autoencoder (IWAE)}

The above ELBO is a lower bound of the true data log-likelihood $\log
p(x)$, hence \cite{DBLP:journals/corr/BurdaGS15} proposed IWAE to
directly estimate the true data log-likelihood with the presence of
the variational model\footnote{The variational model is also referred
  to as the inference model, hence we use them interchangeably.},
namely
\begin{align}
  \log p(x)=\log \mathbb{E}_q\frac{p(x,z)}{q(z|x)}\geq\log\frac{1}{m}\sum_{i=1}^m\frac{p(x,z_i)}{q(z_i|x)}\equiv\mathcal{L}_{\textsc{IWAE}}(x)
\end{align}
where $m$ is the number of importance weighted samples. The above
bound is tighter than the ELBO used in VAE. When trained on the same
network structure as VAE, with the above estimate as training
objective, IWAE achieves considerable improvements over VAE on various
density estimation tasks~\cite{DBLP:journals/corr/BurdaGS15} and
similar idea is also considered in~\cite{DBLP:conf/icml/MnihR16}.

\section{The Proposed Method}
\label{sec:model}

\subsection{Variational Posterior with Auxiliary Variables}

Consider the case of modeling binary data with classic VAE and IWAE,
which typically assumes that a data point is generated from a
multivariate Bernoulli, conditioned on a latent code which is assumed
to be from a Gaussian prior, it's easy to verify that the Gaussian
variational posterior inferred by VAE and IWAE will not match the
non-Gaussian true posterior.

\begin{figure*}[t]
  \begin{subfigure}[t]{0.32\textwidth}
    \centering
    \begin {tikzpicture}[-latex, auto, on grid , semithick , var/.style ={circle ,top color =white , draw, text=black , minimum width =0.5 cm}]
      \node[var] (x) {$\vect z$};
      \node[var] (z) [below=of x] {$\vect x$};
      \node[var] (pz) [right=of x]{$\vect x$};
      \node[var] (px) [below=of pz]{$\vect z$};
      \path (x) edge (z);
      \path (pz) edge (px);
    \end{tikzpicture}
    \caption{Generative model (left) and inference model (right) for VAE}
    \label{fig:vae_model}
  \end{subfigure}
  ~
  \begin{subfigure}[t]{0.32\textwidth}
    \centering
    \begin{tikzpicture}[-latex, auto, on grid , semithick , every node, var/.style ={circle ,top color =white , draw, text=black, minimum width =0.6 cm, inner sep=0pt}]
    \node[var] (x) {$\vect x$};
    \node[var] (tau) [right =of x] {$\vect \tau$};
    \node[var] (z) [below =of tau] {$\vect z$};
    \path (x) edge (tau);
    \path (tau) edge (z);
    \path (x) edge (z);
    \end{tikzpicture}
    \caption{Inference model for AVAE (Generative model is the same as
      in VAE)}
    \label{fig:avae1_model}
  \end{subfigure}
  ~
  \begin{subfigure}[t]{0.32\textwidth}
  \centering
  \begin{tikzpicture}[-latex, auto, on grid , semithick ,every node, var/.style ={circle,radius=1cm,top color =white , draw, text=black, minimum width =0.6 cm, inner sep=0pt}]
    \node[var] (x) {$\vect x$};
    \node[var] (tau1) [right =of x] {$ \tau_1$};
    \node[var] (tau2) [right =of tau1] {$ \tau_2$};
    \node[var] (eps)  [right =of tau2] {$\cdot\cdot\cdot$};
    \node[var] (tauk) [right =of eps] {$ \tau_k$};
    \node[var] (z) [below =of tauk] {$\vect z$};
    \path (x) edge (tau1);
    \path (tau1) edge (tau2);
    \path (tau2) edge (eps);
    \path (eps) edge (tauk);
    \path (tauk) edge (z);
    \path (x) edge [bend right=30](z);
    \path (tau1) edge [bend right=30](z);
    \path (tau2) edge [bend right=15](z);
    \path (eps) edge [bend right=15](z);
  \end{tikzpicture}
  \caption{Inference model for AVAE with $k$ auxiliary variables}
  \label{fig:avaek_model}
  \end{subfigure}
  \caption{Inference models for VAE, AVAE and AVAE with $k$ auxiliary
    random variables (The generative model is fixed as shown in Figure
    \ref{fig:vae_model}). Note that multiple arrows pointing to a node
    indicate one stochastic layer, with the source nodes concatenated
    as input to the stochastic layer and the target node as stochastic
    output. One stochastic layer could consist of multiple
    deterministic layers. (For detailed architecture used in
    experiments, refer to Section \ref{sec:exp}.)}
  \label{fig:inf1}
\end{figure*}
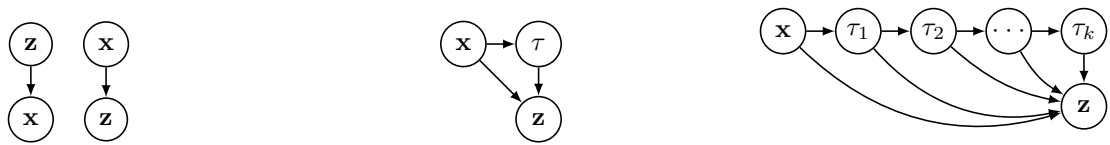

To this end, we propose to introduce an auxiliary random variable
$\tau$ to the inference model of VAE and IWAE. Conditioned on the
input $x$, the inference model equipped with auxiliary variable $\tau$
now defines a joint density over $(\tau,z)$ as
\begin{align}
  q(z,\tau|x)=q(\tau|x)q(z|\tau,x)
\end{align}
where we assume $\tau$ has proper support and both $q(\tau|x)$ and
$q(z|\tau,x)$ can be parameterized. Accordingly the marginal
variational posterior of $z$ given $x$ turns to be
\begin{align}
  q(z|x)&=\int_{\tau}q(z,\tau|x)d\tau=\int_{\tau}q(z|\tau,x)q(\tau|x)d\tau\nonumber\\
  &=\mathbb{E}_{q(\tau|x)}q(z|\tau,x)
\end{align}
which essentially models the posterior $q(z|x)$ as a probabilistic
mixture of different densities $q(z|\tau,x)$ indexed by $\tau$,
together with $q(\tau|x)$ as the mixture weights. This allows complex
and flexible posterior $q(z|x)$ to be constructed, even when both
$q(\tau|x)$ and $q(z|\tau,x)$ are from simple density families. Due to
the presence of auxiliary variables $\tau$, the inference model is
trying to capture more sources of stochasticity than the generative
model, hence we term our approach as Asymmetric Variational
Autoencoder (AVAE). Figure \ref{fig:vae_model} and
\ref{fig:avae1_model} present a comparison of the inference models
between classic VAE and the proposed AVAE.

In the context of of VAE and IWAE, the proposed approach includes two instantiations, AVAE
and IW-AVAE, with loss functions
\begin{align}
  &\mathcal{L}_{\textsc{AVAE}}(x)\equiv\mathbb{E}_{q(z|x)}[\log p(x,z)-\log q(z|x)]\nonumber\\
  =&\mathbb{E}_{q(z|x)}\Big(\log p(x,z)-\log \mathbb{E}_{q(\tau|x)}q(z|\tau,x)\Big)
\end{align}
and 
\begin{align}
  \mathcal{L}_{\textsc{IW-AVAE}}(x)&\equiv \log\mathbb{E}_{q(z|x)}\frac{ p(x,z)}{ q(z|x)}\nonumber\\
  &=\log\mathbb{E}_{q(z|x)} \frac{ p(x,z)}{\mathbb{E}_{q(\tau|x)} q(z|\tau,x)}
\end{align}
respectively.

AVAE enjoys the following properties:
\begin{itemize}
\item \textbf{VAEs are special cases of AVAE}. Conventional
  variational autoencoders can be seen as special cases of AVAE with
  no auxiliary variables $\tau$ assumed;
\item \textbf{No density evaluations for $\tau$ are required}. One key
  advantage brought by the auxiliary variable $\tau$ is that both
  terms inside the inner expectations of $\mathcal{L}_{\textsc{AVAE}}$
  and $\mathcal{L}_{\textsc{IW-AVAE}}$ do not involve $q(\tau|x)$,
  hence no density evaluations are required when Monte Carlo samples
  of $\tau$ are used to optimize the above bounds. 
\item \textbf{Flexible variational posterior.} To fully enrich
  variational model flexibility, we use a neural network $f$ to
  \textit{implicitly} model $q(\tau|x)$ by sampling $\tau$ given $x$
  and a random Gaussian noise vector $\epsilon$ as
  \begin{align}
    \tau=f(x, \epsilon)\text{ with }\epsilon\sim\mathcal{N}(0,I)
  \end{align}
  Due to the flexible representative power of $f$, the implicit
  density $q(\tau|x)$ can be arbitrarily complex. Further we assume
  $q(z|\tau,x)$ to be Gaussian with its mean and variance
  parameterized by neural networks. Since the actual variational
  posterior $q(z|x)=\mathbb{E}_\tau q(z|x,\tau)$, complex posterior
  can be achieved even a simple density family is assumed for
  $q(z|x,\tau)$, due to the possibly flexible family of
  \textit{implicit} density of $q(\tau|x)$ defined by $f(x,
  \epsilon)$.  (Illustration can be found in Section
  \ref{sec:exp_simulation})
\end{itemize}

For completeness, we briefly include that
\begin{proposition} 
Both $\mathcal{L}_{\textsc{AVAE}}(x)$ and
$\mathcal{L}_{\textsc{IW-AVAE}}(x)$ are lower bounds of the true data
log-likelihood, satisfying
$\log p(x)= \mathcal{L}_{\textsc{IW-AVAE}}(x)\geq\mathcal{L}_{\textsc{AVAE}}(x)$.
\end{proposition}

Proof is trivial from Jensen's inequality, hence it's omitted.

\textbf{Remark 1} Though the first equality holds for any choice of
distribution $q(\tau|x)$ (whether $\tau$ depends on $x$ or not), for
practical estimation with Monte Carlo methods, it becomes an
inequality $(\log p(x)\geq \mathcal{\hat L}_{\textsc{IW-AVAE}}(x))$ and the bound tightens as the number of importance samples
is increased~\cite{DBLP:journals/corr/BurdaGS15}. The second inequality
always holds when estimated with Monte Carlo samples.

\textbf{Remark 2} The above bounds are only concerned with one
auxiliary variable $\tau$, in fact $\tau$ can also be a set of
auxiliary variables. Moreover, with the same motivation, we can make
the variational family of AVAE even more flexible by defining a series
of $k$ auxiliary variables, such that
$q(z,\tau_1,...,\tau_k|x)=q(\tau_1|x)q(\tau_2|\tau_1,
x)...q(\tau_k|\tau_{k-1},x)q(z|\tau_1,...,\tau_k,x)$ with sample
generation process for all $\tau$s defined as
\begin{align}
  \tau_1&=f_1(x, \epsilon_1)\nonumber\\
  \tau_{i}&=f_{i}(\tau_{i-1},\epsilon_k) \,\mbox{ for $i=2,3,...,k$}
\end{align}
where all $\epsilon_i$ are random noise vectors and all $f_i$ are neural networks
to be learned. Accordingly, we have
\begin{proposition}
  The AVAE with $k$ auxiliary random variables
  $\{\tau_1,\tau_2,...,\tau_k\}$ is also a lower bound to the true
  log-likelihood, satisfying $\log p(x)=\mathcal{L}_{\textsc{IW-AVAE-}k}\geq
  \mathcal{L}_{\textsc{AVAE-}k}$, where
  \begin{align}
    &\mathcal{L}_{\textsc{AVAE-}k}(x)\equiv
    \mathbb{E}_{q(z|x)} [\log p(x,z)-\log q(z|x)]\nonumber\\
    =&\mathbb{E}_{q(z|x)} \Big(\log p(x,z)
      -\log\mathbb{E}_{q(\tau_1,\tau_2,...,\tau_k|x)}  q(z|\tau_1,...,\tau_k,x)\Big)
  \end{align}
  and
  \begin{align}
    &\mathcal{L}_{\textsc{IW-AVAE-}k}(x)\equiv
    \log\mathbb{E}_{q(z|x)} \frac{ p(x,z)}{ q(z|x)}\nonumber\\
    =&\log\mathbb{E}_{q(z|x)} \frac{ p(x,z)}{\mathbb{E}_{q(\tau_1,\tau_2,...,\tau_k|x)}  q(z|\tau_1,...,\tau_k,x)}
    \label{eq:avae_bound_k}
  \end{align}
\end{proposition}
Figure \ref{fig:avaek_model} illustrates the inference model of an AVAE
with $k$ auxiliary variables. 

\subsection{Learning with Importance Weighted Auxiliary Samples}


For both AVAE and IW-AVAE, we can estimate the corresponding bounds
and its gradients of $\mathcal{L}_{\textsc{AVAE}}$ and
$\mathcal{L}_{\textsc{IW-AVAE}}$ with ancestral sampling from the
model. For example, for AVAE with one auxiliary variable $\tau$, we
estimate
\begin{align}
  \mathcal{\hat L}_{\textsc{AVAE}}(x)=\frac{1}{m}\sum_{i=1}^{m}\left(\log p(x, z_i)-\log \frac{1}{n}\sum_{j=1}^nq(z_i|\tau_j, x)\right)
\end{align}
and
\begin{align}
  \mathcal{\hat L}_{\textsc{IW-AVAE}}(x)=\log\frac{1}{m}\sum_{i=1}^{m}\frac{ p(x, z_i)}{\frac{1}{n}\sum_{j=1}^nq(z_i|\tau_j, x)}
\end{align}
where $n$ is the number of $\tau$s sampled from the current
$q(\tau|x)$ and $m$ is the number of $z$s sampled from the
\textit{implicit} conditional $q(z|x)$, which is by definition
achieved by first sampling from $q(\tau|x)$ and subsequently sampling
from $q(z|\tau,x)$. The parameters of both the inference model and
generative model are jointly learned by maximizing the above
bounds. Besides back propagation through the stochastic variable $z$
(typically assumed to be a Gaussian for continuous latent variables)
is possible through the re-parameterization trick, and it is naturally
also true for all the auxiliary variables $\tau$ since they are
constructed in a generative manner.

The term $\frac{1}{n}\sum_{j=1}^n q(z_i|\tau_j,x)$ essentially is an
$n$-sample importance weighted estimate of $q(z|x)=\mathbb{E}_\tau
q(z|\tau,x)$, hence it is reasonable to believe that more samples of
$\tau$ will lead to less noisy estimate of $q(\tau|x)$ and thus a more
accurate inference model $q$. It's worth pointing out for AVAE that
additional samples of $\tau$ comes almost at no cost when multiple
samples of $z$ are generated $(m>1)$ to optimize
$\mathcal{L}{_\textsc{AVAE}}$ and $\mathcal{L}{_\textsc{IW-AVAE}}$,
since sampling a $z$ from the inference model will also generate
intermediate samples of $\tau$, thus we can always reuse those samples
of $\tau$ to estimate $q(z|x)=\mathbb{E}_\tau q(z|\tau,x)$. For this
purpose, in our experiments we always assume $n=m$ so that no separate
process of sampling $\tau$ is needed in estimating the lower
bounds. This also ensures that the forward pass and backward pass time
complexity of the inference model are the same as conventional VAE and
IWAE. In fact, as we will show in all our empirical evaluations that
if $n=1$ AVAE performs similarly to VAE and while $n>1$ IW-AVAE always
outperforms IWAE, i.e., its counterpart with no auxiliary variables.

\section{Connection to Related Methods}
\label{sec:related}

Before we proceed to the experimental evaluations of the proposed
methods, we highlight the relations of AVAE to other similar methods.

\subsection{Other methods with auxiliary variables}

Relation to Hierarchical Variational Models
(HVM)~\cite{DBLP:conf/icml/RanganathTB16} and Auxiliary Deep
Generative Models (ADGM)~\cite{DBLP:conf/icml/MaaloeSSW16} are two
closely related variational methods with auxiliary variables. HVM also
considers enriching the variational model family by placing a prior
over the latent variable for the variational distribution
$q(z|x)$. While ADGM takes another way to this goal, by placing a
prior over the auxiliary variable on the generative model, which in
some cases will keep the marginal generative distribution of the data
invariant. It has been shown that HVM and ADGM are mathematically
equivalent by~\cite{brummer2016note}.

However, our proposed method doesn't add any prior on the generative
model and thus doesn't change the structure of the generative
model. We emphasize that our proposed method makes the least
assumption about the generative model and that the proposal in our
method is orthogonal to related methods, thus it can can be integrated
with previous methods with auxiliary variables to further boost the
performance on accurate posterior approximation and generative
modeling.

\subsection{Adversarial learning based inference models}

Adversarial learning based inference models, such as Adversarial
Autoencoders~\cite{makhzani2015adversarial}, Adversarial Variational
Bayes~\cite{mescheder2017adversarial}, and Adversarially Learned
Inference~\cite{dumoulin2016adversarially}, aim to maximize the ELBO
without any variational likelihood evaluations at all. It can be shown
that for the above adversarial learning based models, when the
discriminator is trained to its optimum, the model is equivalent to
optimizing the ELBO. However, due to the minimax game involved in the
adversarial setting, practically at any moment it is not guaranteed
that they are optimizing a lower bound of the true data likelihood,
thus no maximum likelihood learning interpretation can be
provided. Instead in our proposed framework, we don't require
variational density evaluations for the flexible auxiliary variables,
while still maintaining the maximum likelihood interpretation.

\section{Experiments}
\label{sec:exp}

\subsection{Flexible Variational Family of AVAE}
\label{sec:exp_simulation}

To test the effect of adding auxiliary variables to the inference
model, we parameterize two unnormalized 2D target densities $p(\vect
z)\propto \exp(U(\vect z))$\footnote{Sample densities originate
  from~\cite{DBLP:conf/icml/RezendeM15}} with
\begin{align}
  U_1(\vect z)=&\frac{1}{2}\left(\frac{\|\vect z\|-2}{4}\right)^2\nonumber\\
  &-\log\left(e^{-\frac{1}{2}\left[\frac{\vect z_1-2}{0.6}\right]^2}+e^{-\frac{1}{2}\left[\frac{\vect z_1+2}{0.6}\right]^2}\right)\nonumber\\
\text{and }U_2(\vect  z)&=\frac{1}{2}\left[\frac{\vect z_2-w_1(\vect z)}{0.4}\right]^2\,\text{ where }w_1(\vect z)=\sin\left(\frac{\pi\vect
  z_1}{2}\right)\nonumber
\end{align}
We construct inference model\footnote{Inference model of VAE defines a
  conditional variational posterior $q(z|x)$, to match the target
  density $p(z)$ which is independent of $x$, we set $x$ to be
  fixed. In this synthetic example, $x$ is set to be an all one vector
  of dimension 10.} to approximate the target density by minimizing
the KL divergence
\begin{align}
  \text{KL}(q(z)\|p(z))&=\mathbb{E}_{z\sim q(z)}\big(\log q(z)-\log p(z)\big)\nonumber\\
  &=\mathbb{E}_{z\sim q(z)}\big(\log \mathbb{E}_\tau q(z|\tau)-\log p(z)\big)
\end{align}
\begin{figure}[!htb]
  \centering
  \begin{subfigure}[t]{0.15\textwidth}
    \includegraphics[width=\textwidth]{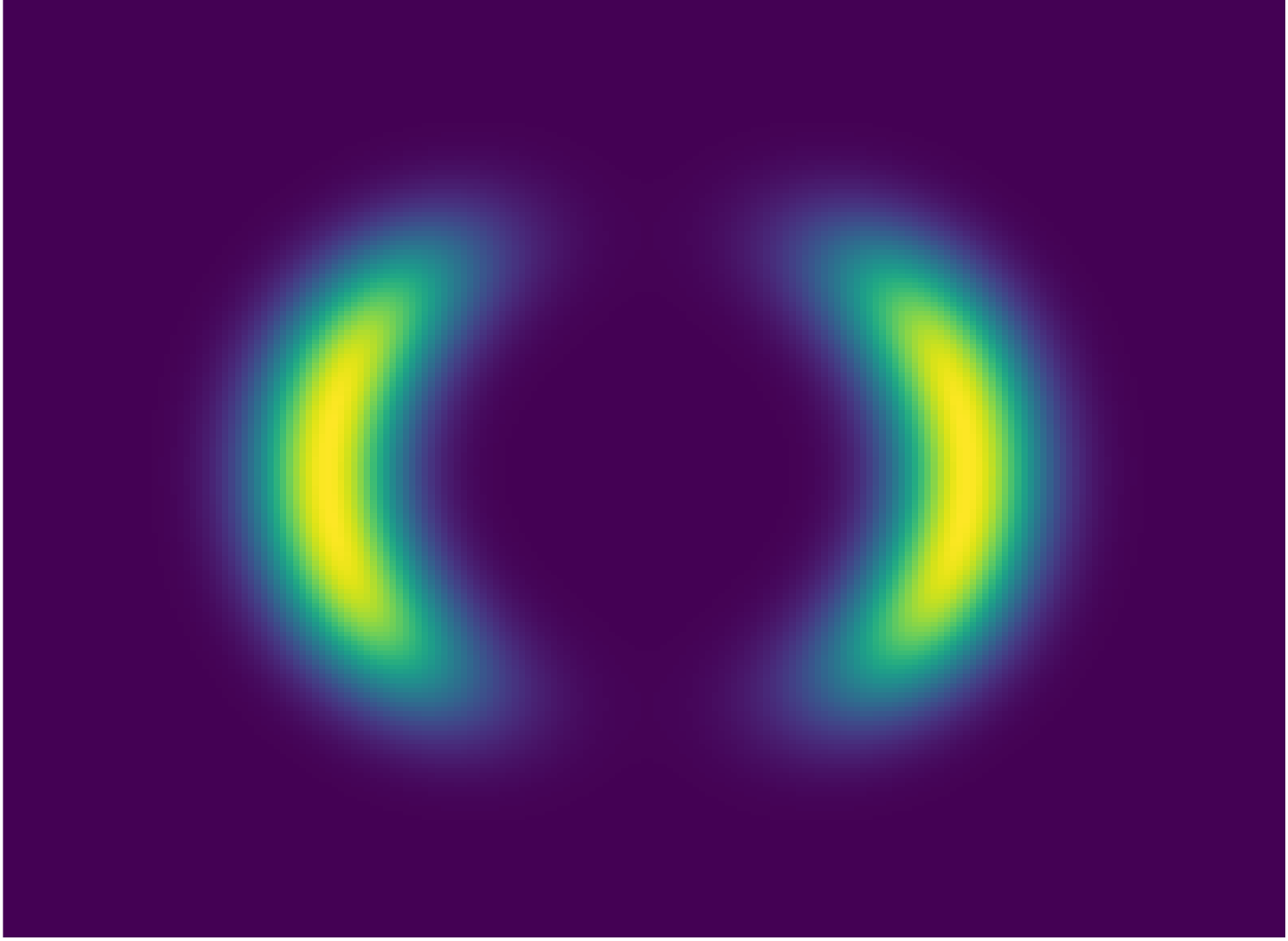}
  \end{subfigure}
  ~
  \begin{subfigure}[t]{0.15\textwidth}
    \includegraphics[width=\textwidth]{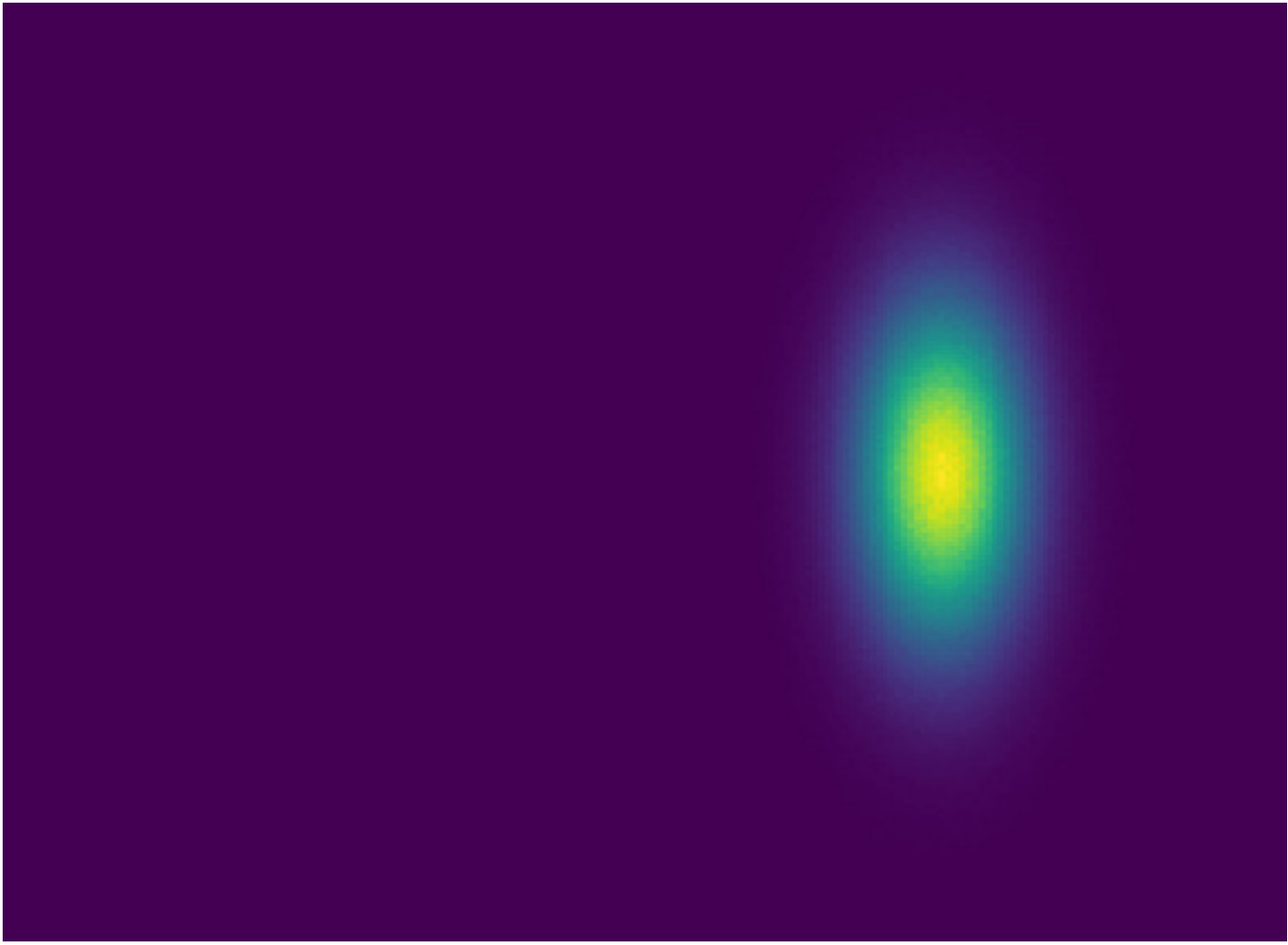}
  \end{subfigure}
  ~
  \begin{subfigure}[t]{0.15\textwidth}
    \includegraphics[width=\textwidth]{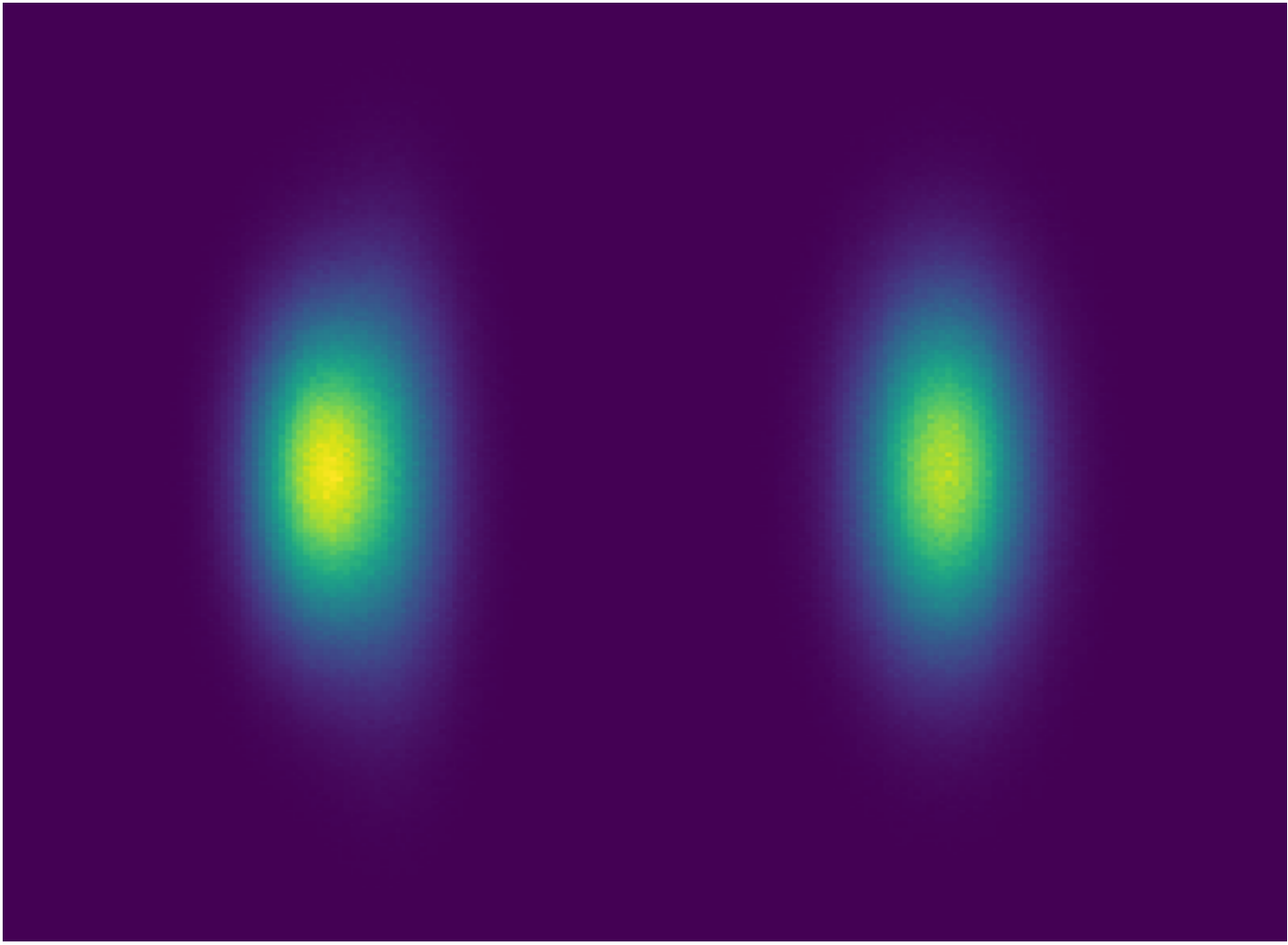}
  \end{subfigure}
  \\
    \begin{subfigure}[t]{0.15\textwidth}
      \includegraphics[width=\textwidth]{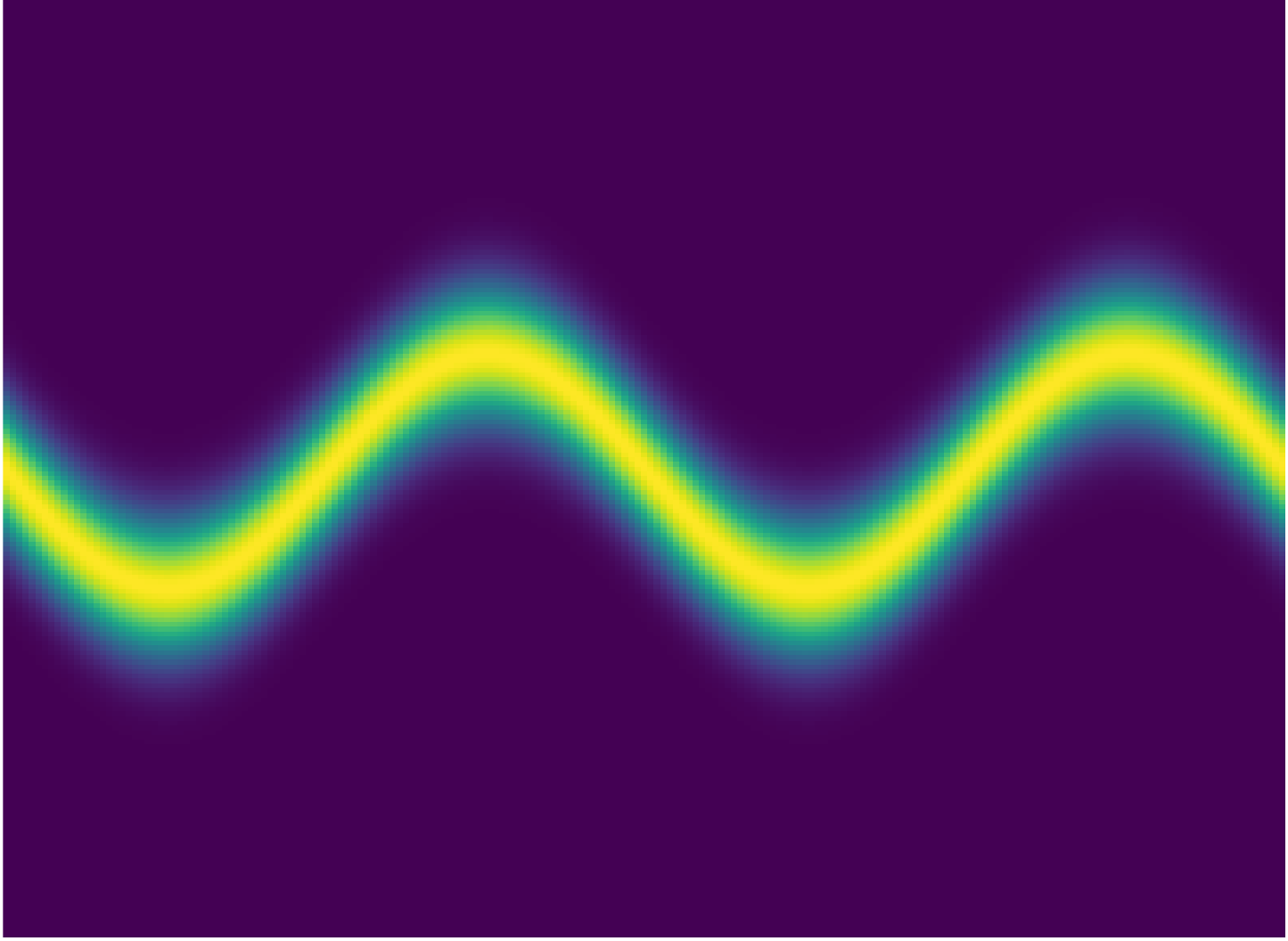}
      \caption{}
    \end{subfigure}
  ~
  \begin{subfigure}[t]{0.15\textwidth}
    \includegraphics[width=\textwidth]{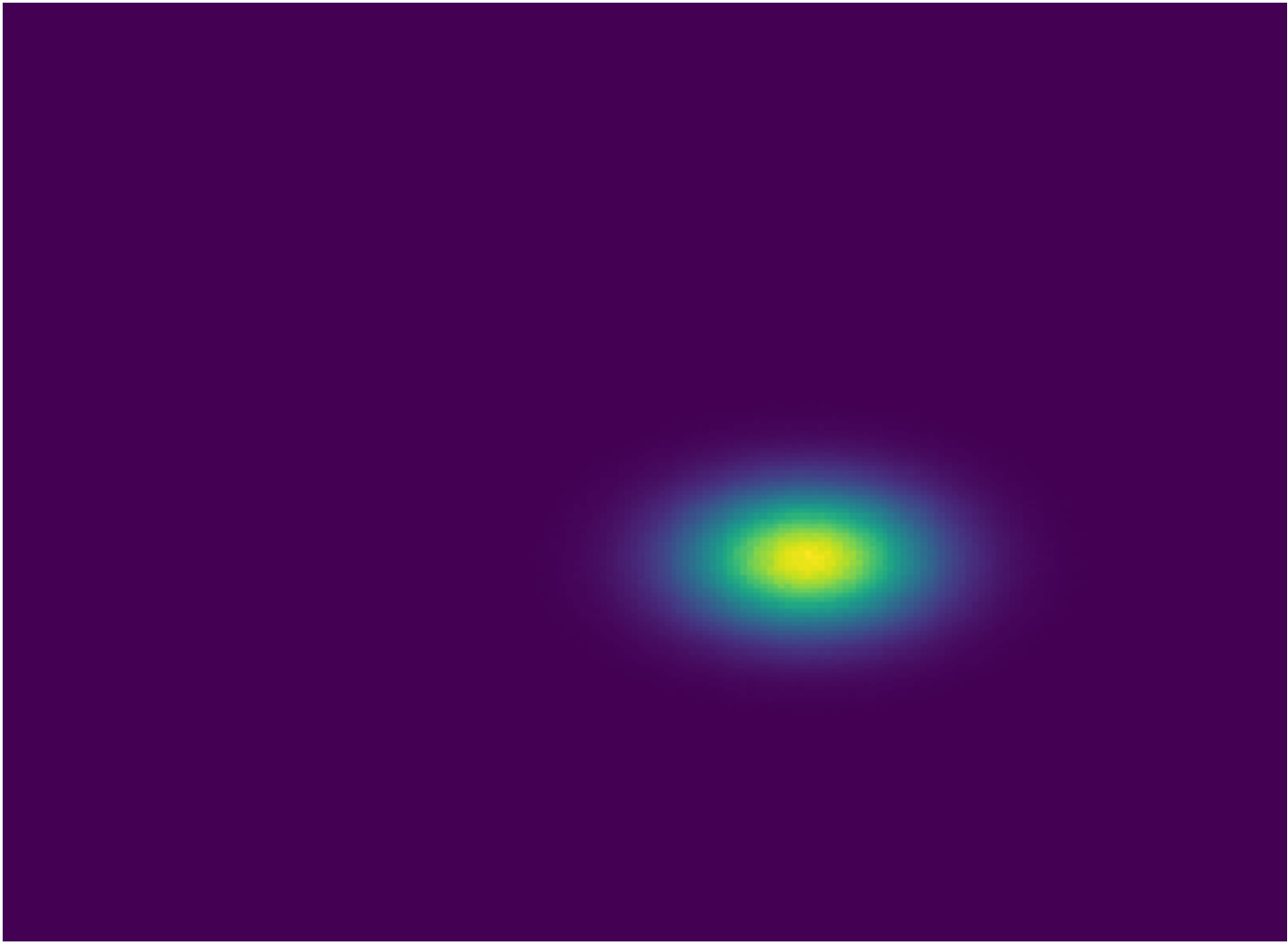}
    \caption{}
    \label{fig:flow_vae}
  \end{subfigure}
  ~
  \begin{subfigure}[t]{0.15\textwidth}
    \includegraphics[width=\textwidth]{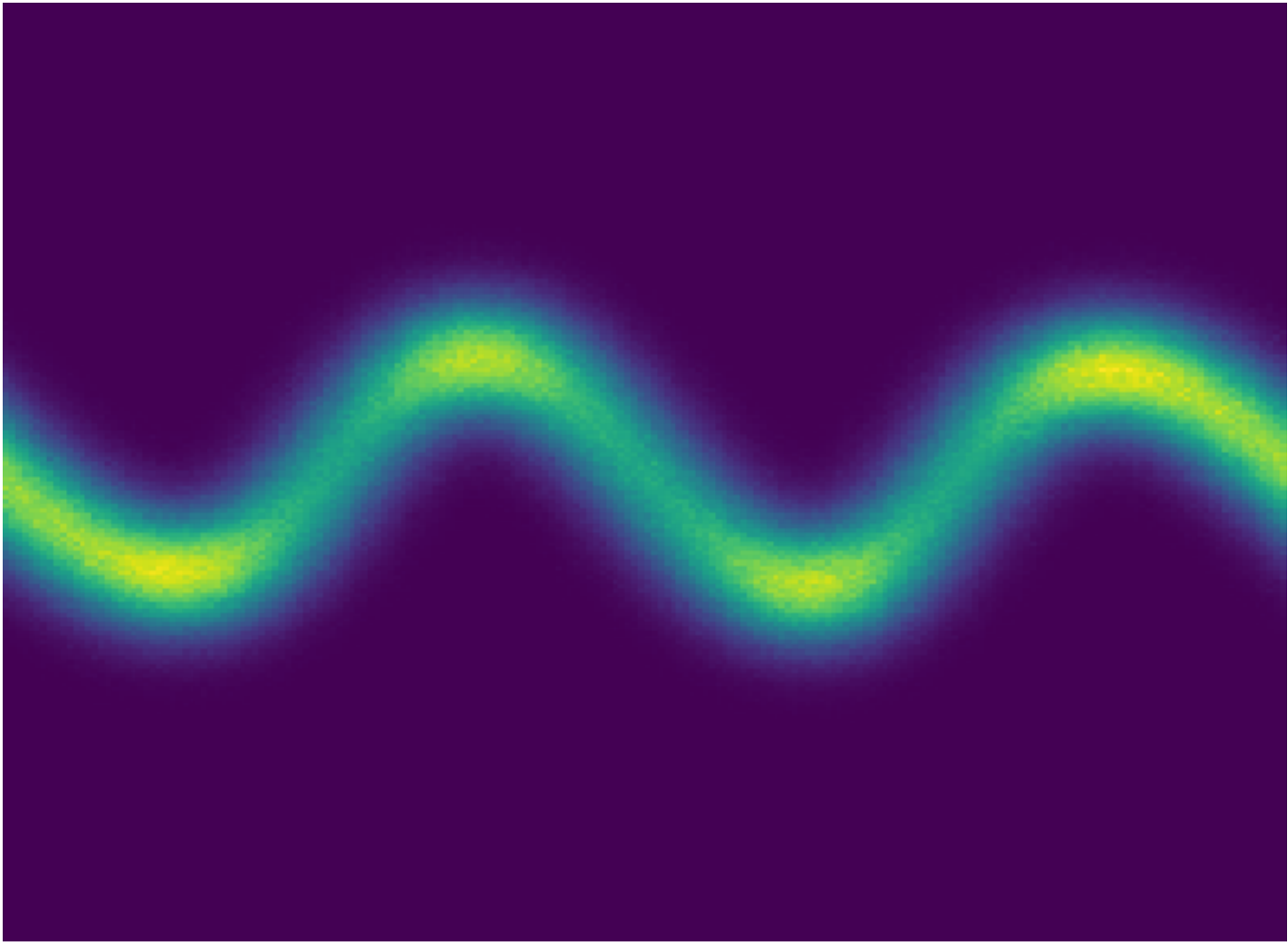}
    \caption{}
    \label{fig:flow_avae}
  \end{subfigure}
  \caption{(a) True density; (b) Density learned by VAE; (c) Density
    learned by AVAE.}
  \label{fig:flow}
  \vspace{-0.15in}
\end{figure}
Figure \ref{fig:flow} illustrates the target densities as well as the
ones learned by VAE and AVAE, respectively. It's unsurprising to see
that standard VAE with Gaussian stochastic layer as its inference
model will only be able to produce Gaussian density estimates (Figure
\ref{fig:flow_vae}). While with the help of introduced auxiliary
random variables, AVAE is able to match the non-Gaussian target
densities (Figure \ref{fig:flow_avae}), even the last stochastic layer
of the inference model, i.e., $q(z|\tau)$, is also Gaussian.

\subsection{Handwritten Digits and Characters}

To test AVAE for variational inference we use standard benchmark
datasets
MNIST\footnote{\url{http://www.cs.toronto.edu/~larocheh/public/datasets/binarized_mnist/}}
and
OMNIGLOT\footnote{\url{https://github.com/yburda/iwae/raw/master/datasets/OMNIGLOT/chardata.mat}}~\cite{DBLP:conf/nips/LakeST13}. Our
method is general and can be applied to any formulation of the
generative model $p_\theta(x,z)$. For simplicity and fair comparison,
in this paper we focus on $p_\theta(x,z)$ defined by stochastic neural
networks, i.e., a family of generative models with their parameters
defined by neural networks. Specifically, we consider the following
two types of generative models:
\begin{enumerate}
\item[$G_1:$] $p_\theta(x,z)=p_\theta(z)p_\theta(x|z)$ with single
  Gaussian stochastic layer for $z$ with 50 units. In between the
  latent variable $z$ and observation $x$ there are two deterministic
  layers, each with 200 units;
\item[$G_2:$] $p_\theta(x,z_1,
  z_2)=p_\theta(z_1)p_\theta(z_2|z_1)p_\theta(x|z_2)$ with two
  Gaussian stochastic layers for $z_1$ and $z_2$ with 50 and 100
  units, respectively. Two deterministic layers with 200 units connect
  the observation $x$ and latent variable $z_2$, and two deterministic
  layers with 100 units connect $z_2$ and $z_1$.
\end{enumerate}
A Gaussian stochastic layer consists of two fully connected linear
layers, with one outputting the mean and the other outputting the
logarithm of diagonal covariance. All other deterministic layers are
fully connected with tanh nonlinearity. The same network architectures
for both $G_1$ and $G_2$ are also used in
~\cite{DBLP:journals/corr/BurdaGS15}

For $G_1$, an inference network with the following architecture is
used by AVAE with $k$ auxiliary variables
\begin{align}
  &\tau_{i}=f_i(\tau_{i-1}\|\epsilon_i)\,\mbox{ where }\epsilon_i\sim\mathcal{N}(0, I)\mbox{ for $i=1,2,...,k$}\nonumber\\
  &q(z|x, \tau_1,...,\tau_k)=\mathcal{N}\Big(\mu(x\|\tau_1\|...\|\tau_k),\mbox{diag}\big(\sigma(x\|\tau_1\|...\|\tau_k)\Big)\nonumber
\end{align}
where $\tau_0$ is defined as input $x$, all $f_i$ are implemented as
fully connected layers with tanh nonlinearity and $\|$ denotes the
concatenation operator. All noise vectors $\epsilon$s are set to be of
50 dimensions, and all other variables have the corresponding
dimensions in the generative model. Inference network used for $G_2$
is the same, except that the Gaussian stochastic layer is defined for
$z_2$. An additional Gaussian stochastic layer for $z_1$ is defined
with $z_2$ as input, where the dimensions of variables aligned to
those in the generative model $G_2$.  Further, Bernoulli observation
models are assumed for both MNIST and OMNIGLOT. For MNIST, we employ
the static binarization strategy as
in~\cite{DBLP:journals/jmlr/LarochelleM11} while dynamic binarization
is employed for OMNIGLOT.

Our baseline models include VAE and IWAE. Since our proposed method
involves adding more layers to the inference network, we also include
another enhanced version of VAE with more deterministic layers added
to its inference network, which we term as VAE+\footnote{VAE+ is a
  restricted version of AVAE with all the noise vectors $\epsilon$s
  set to be constantly 0, but with the additional layers for $f$s
  retained.} and its importance sample weighted variant IWAE+. To
eliminate discrepancies in implementation details of the models
reported in the literature, we implement all models and carry out the
experiments under the same setting: All models are implemented in
PyTorch\footnote{\url{http://pytorch.org/}} and parameters of all
models are optimized with Adam~\cite{DBLP:journals/corr/KingmaB14} for
2000 epochs, with an initial learning rate of 0.001, cosine annealing
for learning rate decay~\cite{loshchilov2016sgdr}, exponential decay
rates for the 1st and 2nd moments at 0.9 and 0.999,
respectively. Batch normalization~\cite{DBLP:conf/icml/IoffeS15} is
applied to all fully connected layers, except for the final output
layer for the generative model, as it has been shown to improve
learning for neural stochastic
models~\cite{DBLP:conf/nips/SonderbyRMSW16}. Linear annealing of the
KL divergence term between the variational posterior and the prior in
all the loss functions from 0 to 1 is adopted for the first 200
epochs, as it has been shown to help training stochastic neural
networks with multiple layers of latent
variables~\cite{DBLP:conf/nips/SonderbyRMSW16}. Code to reproduce all
reported results will be made publicly available.

\subsubsection{Generative Density Estimation}

\begin{table*}[t]
  \centering
  \caption{MNIST and OMNIGLOT test set NLL with generative models
    $G_1$ and $G_2$ (\textit{Lower is better; for VAE+, $k$ is the
      number of additional layers added and for AVAE it is the number
      of auxiliary variables added. For each column, the best result
      for each $k$ of both type of models (VAE based and IWAE based)
      are printed in bold. )}}
  \begin{tabular}{lrrrr}
    \toprule
    & \multicolumn{2}{c}{MNIST}&\multicolumn{2}{c}{OMNIGLOT}\\
    Models & $-\log p(x)$ on $G_1$  & $-\log p(x)$ on $G_2$ & $-\log p(x)$ on $G_1$  & $-\log p(x)$ on $G_2$ \\
    \midrule
    VAE ~\cite{DBLP:journals/corr/BurdaGS15}       & $88.37$ & $85.66$ & $108.22 $ & $106.09$\\

    \midrule
    VAE+ $(k=1)$ & $88.20$ & $ 85.41$ & $108.30$ & $ 106.30$ \\
    VAE+ $(k=4)$ & $88.08$ & $ 85.26$ & $108.31$ & $ 106.48$ \\ 
    VAE+ $(k=8)$ & $87.98$ & $ 85.16$ & $108.31$ & $ 106.05$ \\
    \midrule
    AVAE $(k=1)$ & $88.20$ & $ 85.52$ & $108.27$ & $ 106.59 $ \\
    AVAE $(k=4)$ & $88.18$ & $ 85.36$ & $108.21$ & $ 106.43 $ \\
    AVAE $(k=8)$ & $88.23$ & $ 85.33$ & $108.20$ & $ 106.49$ \\
    \midrule
    AVAE $(k=1,m=50)$ & $\vect{87.21}$ & $\vect{84.57}$ &$\vect{106.89}$  & $\vect{104.59}$\\
    AVAE $(k=4,m=50)$ & $\vect{86.98}$ & $\vect{84.39}$ &$\vect{106.50}$ & $\vect{104.76}$\\ 
    AVAE $(k=8,m=50)$ & $\vect{86.89}$ & $\vect{84.36} $ &$\vect{106.51}$ & $\vect{104.67}$\\

    \toprule
    
    Models (Importance weighted) & & & &\\
    \midrule
    IWAE $(m=50)$~\cite{DBLP:journals/corr/BurdaGS15} & $86.90$ & $84.26$ & $106.08$ & $104.14$ \\
    \midrule
    IW-AVAE $(k=1,m=5)$ & $86.86$ & $ 84.47$ & $106.80$ & $104.67$ \\
    IW-AVAE $(k=4,m=5)$ & $86.57$ & $ 84.55$ & $106.93$ & $104.87$ \\ 
    IW-AVAE $(k=8,m=5)$ & $86.67$ & $ 84.44$ & $106.57$ & $105.06$\\
    
    \midrule

    IWAE+ $(k=1, m=50)$ & $86.70$ & $84.28$  & $105.83$ & $\vect{103.79}$\\
    IWAE+ $(k=4, m=50)$ & $86.31$ & $83.92$ & $105.81$  &$\vect{103.71}$\\ 
    IWAE+ $(k=8, m=50)$ & $86.40$ & $84.06$ & $105.73$  &$\vect{103.77}$\\
        \midrule
    IW-AVAE $(k=1,m=50)$ & $\vect{86.08}$ & $ \vect{84.19} $& $\vect{105.49}$& $103.84$\\
    IW-AVAE $(k=4,m=50)$ & $\vect{86.02}$ & $ \vect{84.05} $& $\vect{105.53}$& $103.89$\\
    IW-AVAE $(k=8,m=50)$ & $\vect{85.89} $ & $ \vect{ 83.77} $& $\vect{105.39}$ & $103.97$\\
    \bottomrule
  \end{tabular}
  \label{tab:mnist}
\end{table*}

For both MNIST and OMNIGLOT, all models are trained and tuned on the
training and validation sets, and estimated log-likelihood on the test
set with 128 importance weighted samples are reported. Table
\ref{tab:mnist} presents the performance of all models with for both
$G_1$ and $G_2$.

Firstly, VAE+ achieves slightly higher log-likelihood estimates than
vanilla VAE due to the additional layers added in the inference
network, implying that a better Gaussian posterior approximation is
learned. Second, AVAE achieves lower NLL estimates than VAE+, more so
with increasingly more samples from auxiliary variables (i.e., larger
$m$), which confirms our expectation that: a) adding auxiliary
variables to the inference network leads to a richer family of
variational distributions; b) more samples of auxiliary variables
yield a more accurate estimate of variational posterior
$q(z|x)$. Lastly, with more importance weighted samples from both
$\tau$ and $z$, i.e., IW-AVAE variants, the best data density
estimates are achieved. Overall, on MNIST AVAE outperforms VAE by 1.5
nats on $G_1$ and 1.3 nats on $G_2$; IW-AVAE outperforms IWAE by about
1.0 nat on $G_1$ and 0.5 nats on $G_2$. Similar trends can be observed
on OMNIGLOT, with AVAE and IW-AVAE outperforming conventional VAE and
IWAE in all cases, except for $G_2$ IWAE+ slightly outperforms
IW-AVAE.

Compared with previous methods with similar settings, IW-AVAE achieves
a best NLL of 83.77, significantly better than 85.10 achieved by
Normalizing Flow~\cite{DBLP:conf/icml/RezendeM15}. Best density
modeling with generative modeling on statically binarized MNIST is
achieved by Pixel RNN~\cite{oord2016pixel,salimans2017pixelcnnpp} with
autoregressive models and Inverse Autoregressive
Flows~\cite{DBLP:conf/nips/KingmaSJCCSW16} with latent variable
models, however it's worth noting that much more sophisticated
generative models are adopted in those methods and that AVAE enhances
standard VAE by focusing on enriching inference model flexibility,
which pursues an orthogonal direction for improvements. Therefore,
AVAE can be integrated with above-mentioned methods to further improve
performance on latent generative modeling.

\subsubsection{Latent Code Visualization}
\begin{figure}[!htb]
  \vspace{-0.1in}
  \centering
  \includegraphics[height=0.23\linewidth]{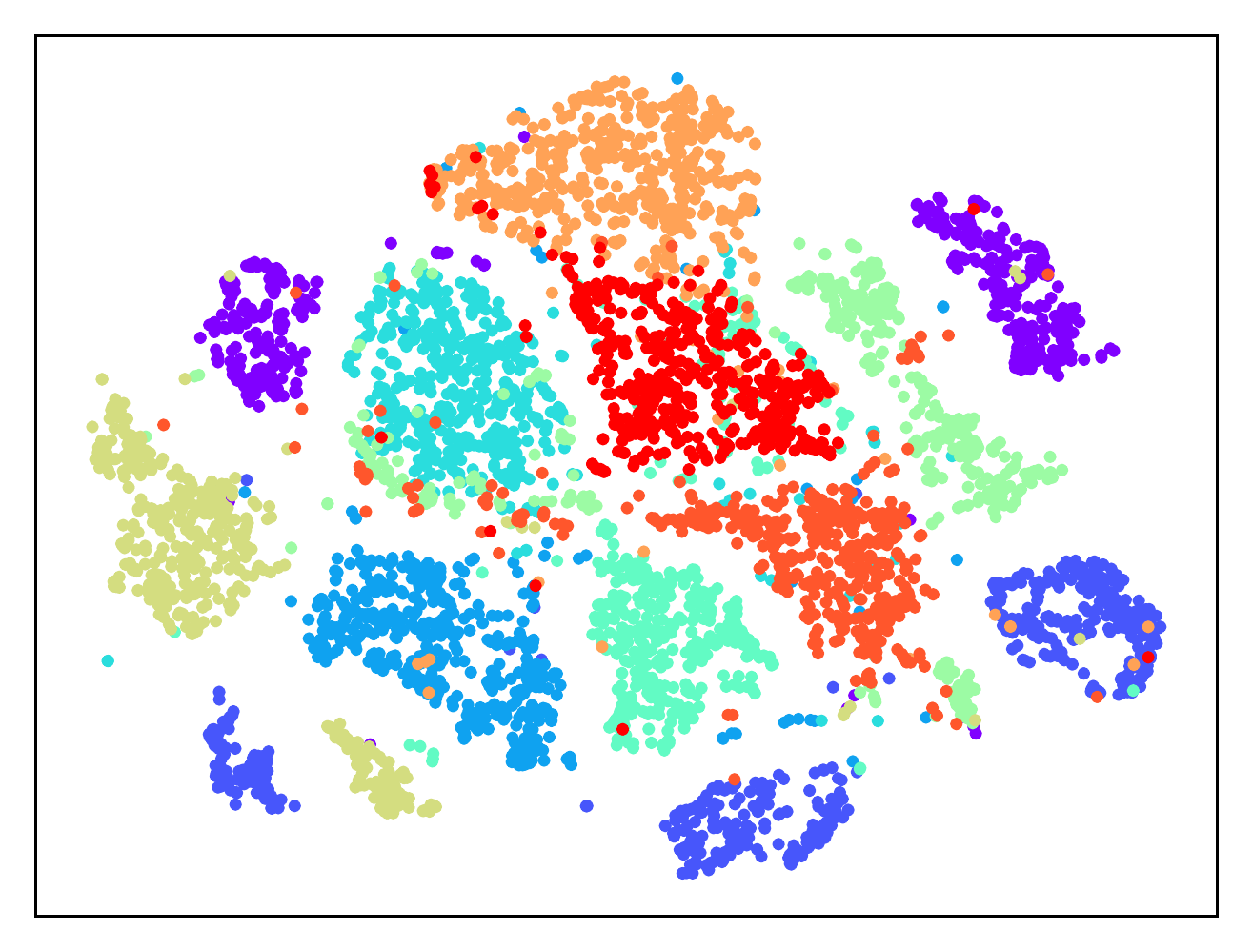}
  \includegraphics[height=0.23\linewidth]{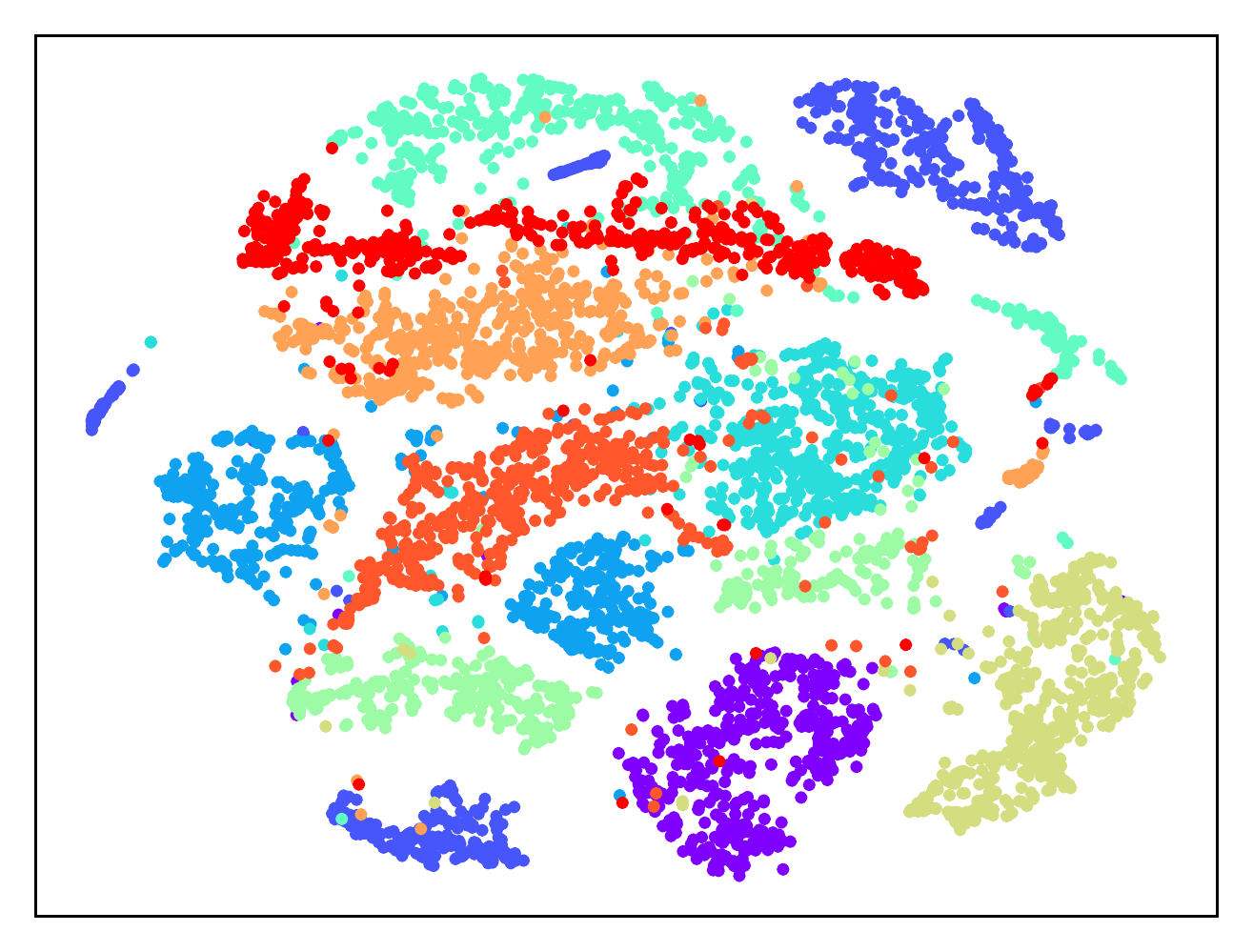}
  \includegraphics[height=0.23\linewidth]{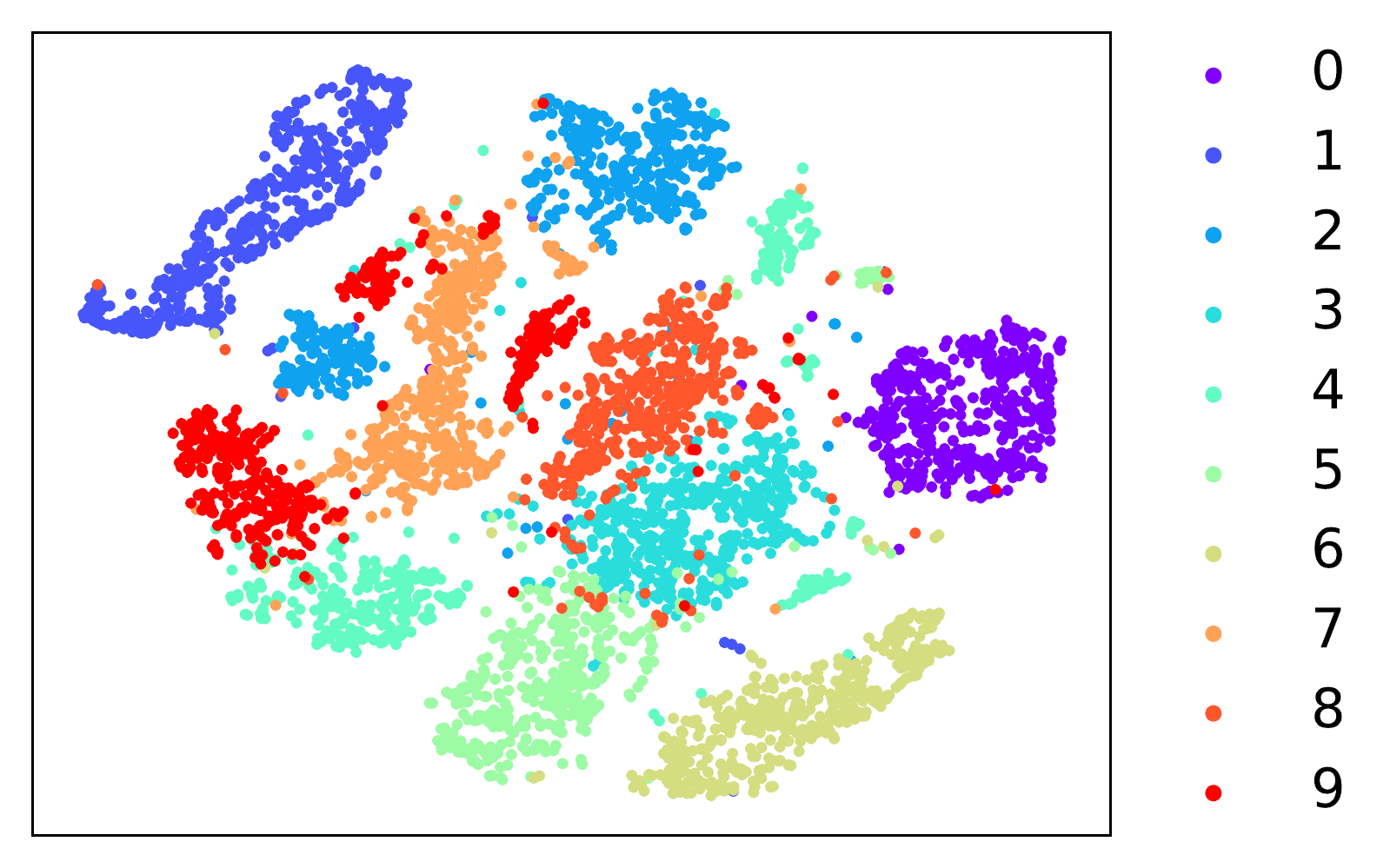}
  \caption{\textbf{Left:} VAE, \textbf{Middle:} VAE+,
    \textbf{Right:}AVAE. Visualization of inferred latent codes for
    5000 MNIST digits in the test set (best viewed in color)}
  \label{fig:tsne}
  \vspace{-0.2in}
\end{figure}

We visualize the inferred latent codes $z$ of digits in the MNIST test
set with respect to their true class labels in Figure \ref{fig:tsne}
from different models with tSNE~\cite{maaten2008visualizing}.
We observe that on generative model $G_2$, all three models are able
to infer latent codes of the digits consistent with their true
classes. However, VAE and VAE+ still shows disconnected cluster of
latent codes from the same class (both class 0 and 1) and latent code
overlapping from different classes (class 3 and 5), while AVAE outputs
clear separable latent codes for different classes (notably for class
0,1,5,6,7).


\subsubsection{Reconstruction and Generated Samples}

\begin{figure}[!htb]
  \vspace{-0.15in}
  \centering
  \begin{subfigure}[t]{0.15\textwidth}
    \includegraphics[width=\textwidth]{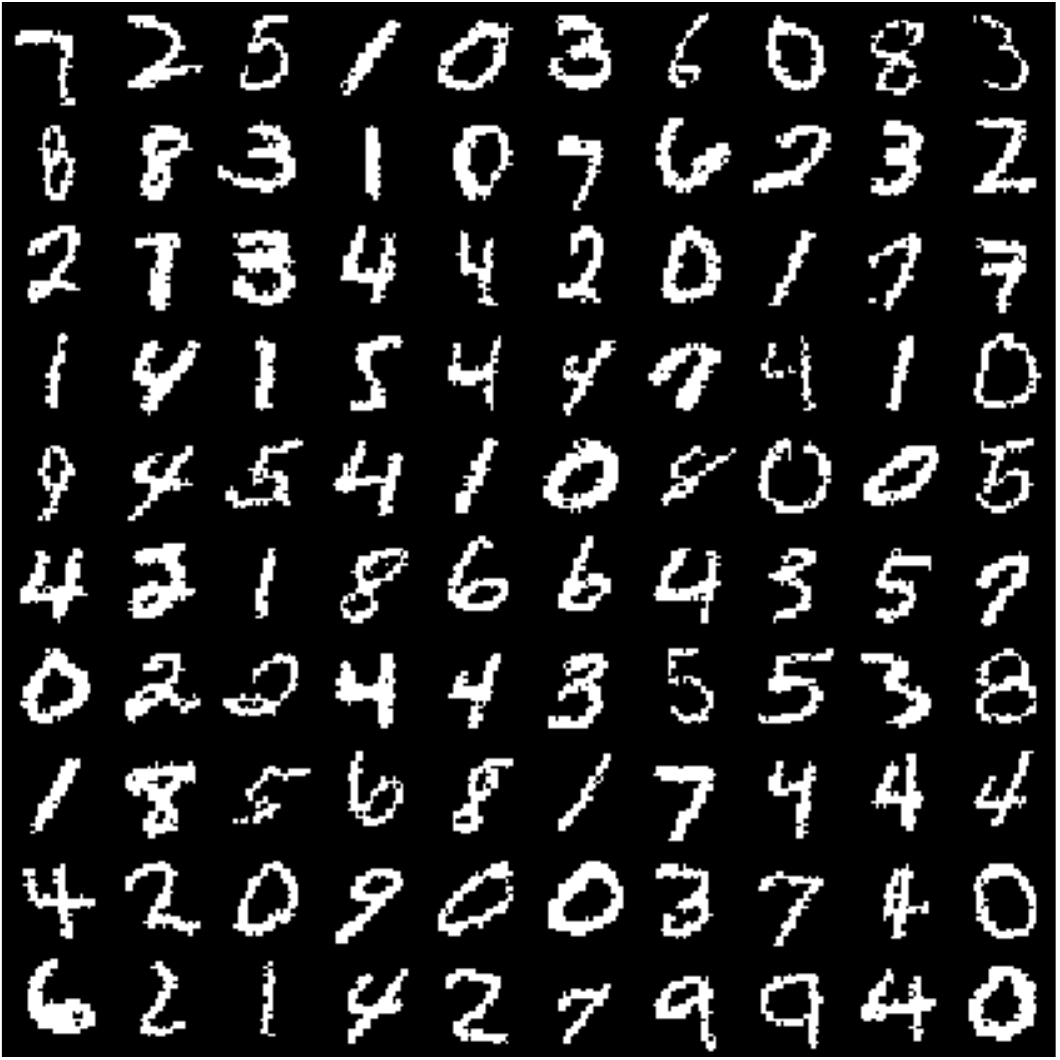}
  \end{subfigure}
  \begin{subfigure}[t]{0.15\textwidth}
    \includegraphics[width=\textwidth]{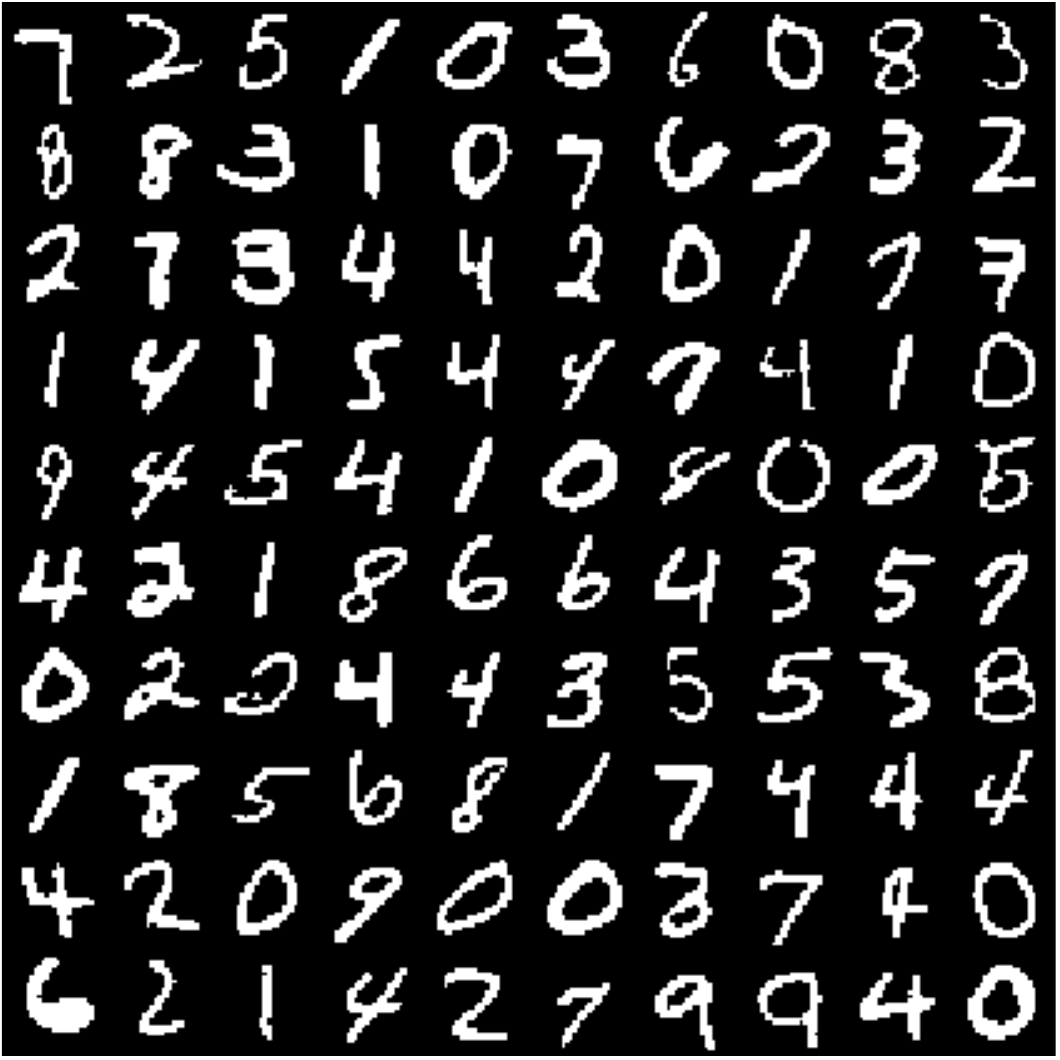}
  \end{subfigure}
  \begin{subfigure}[t]{0.15\textwidth}
    \includegraphics[width=\textwidth]{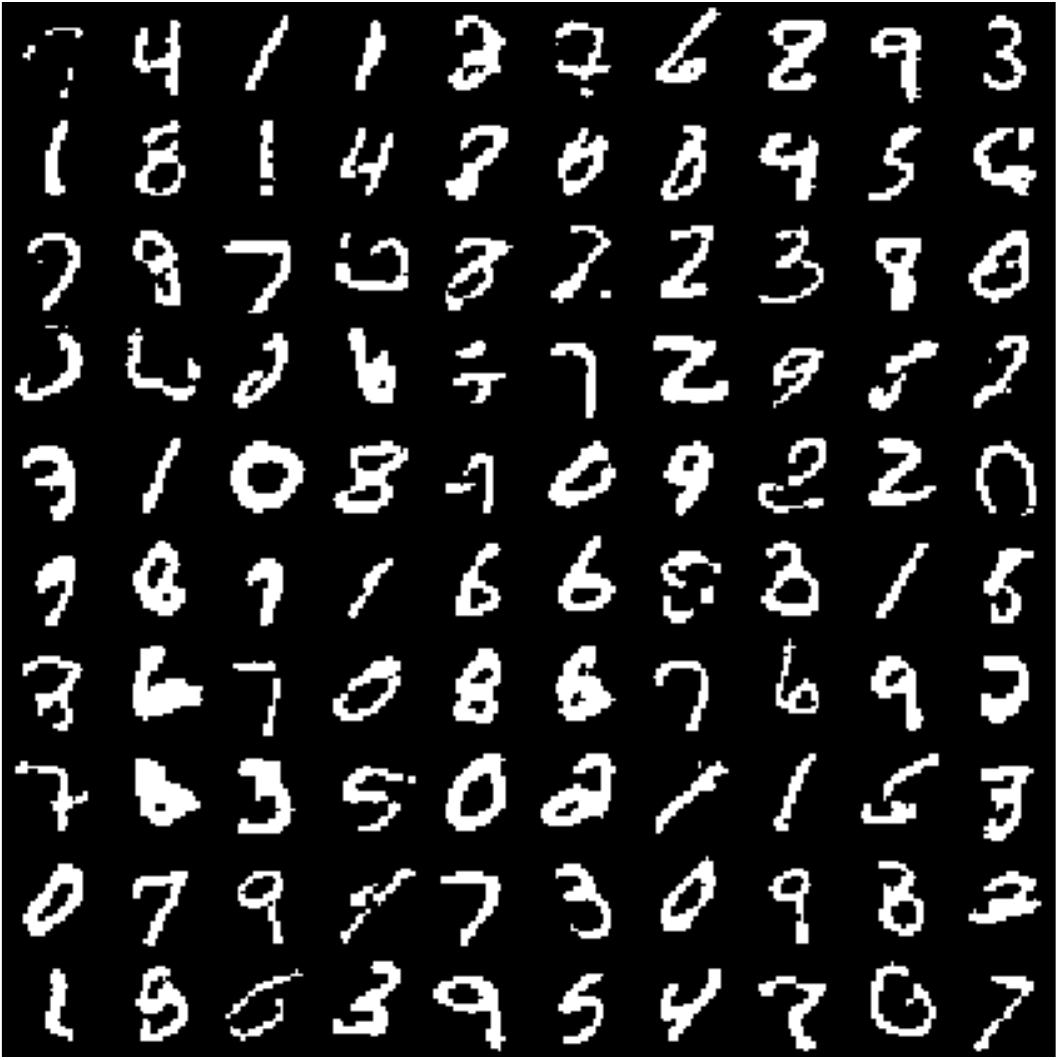}
  \end{subfigure}\\
    \begin{subfigure}[t]{0.15\textwidth}
    \includegraphics[width=\textwidth]{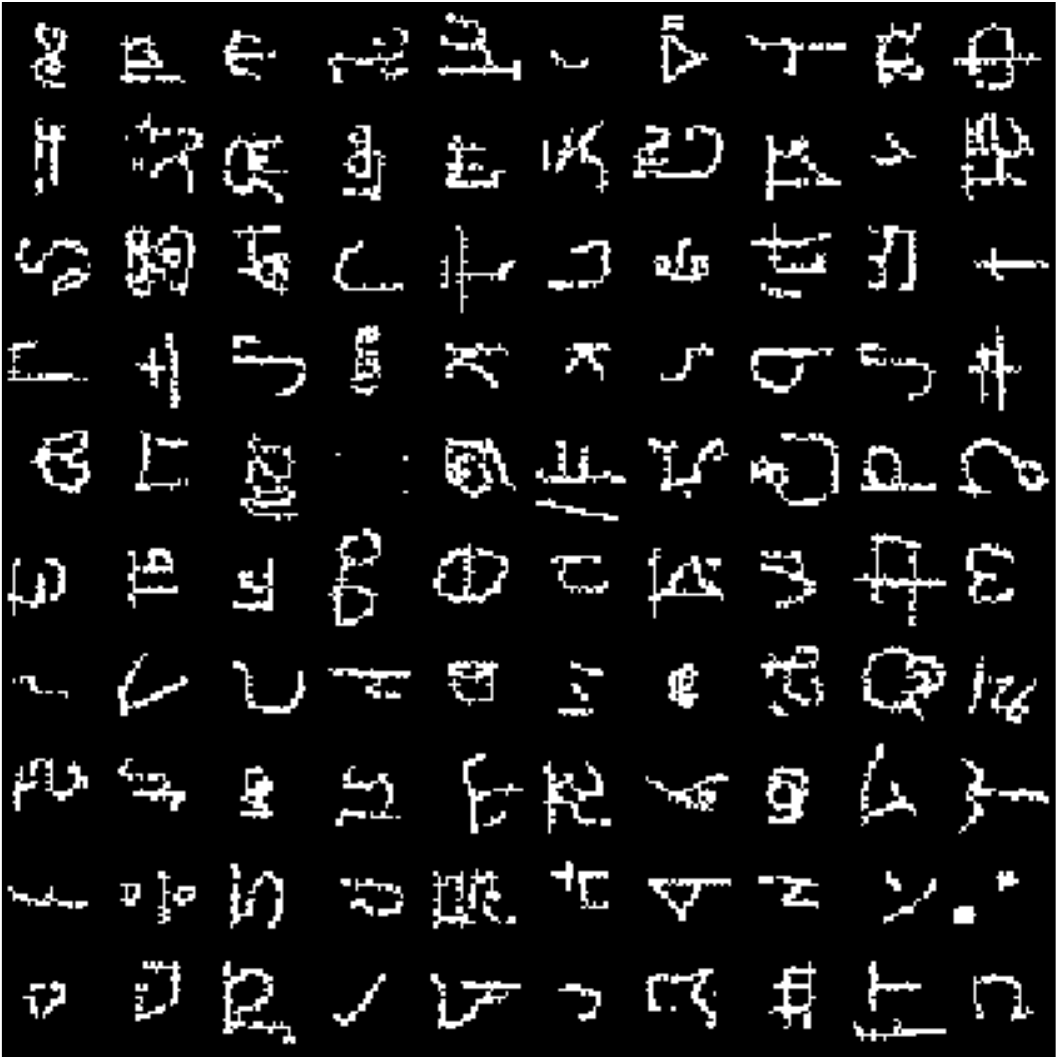}
    \caption{Data}
  \end{subfigure}
  \begin{subfigure}[t]{0.15\textwidth}
    \includegraphics[width=\textwidth]{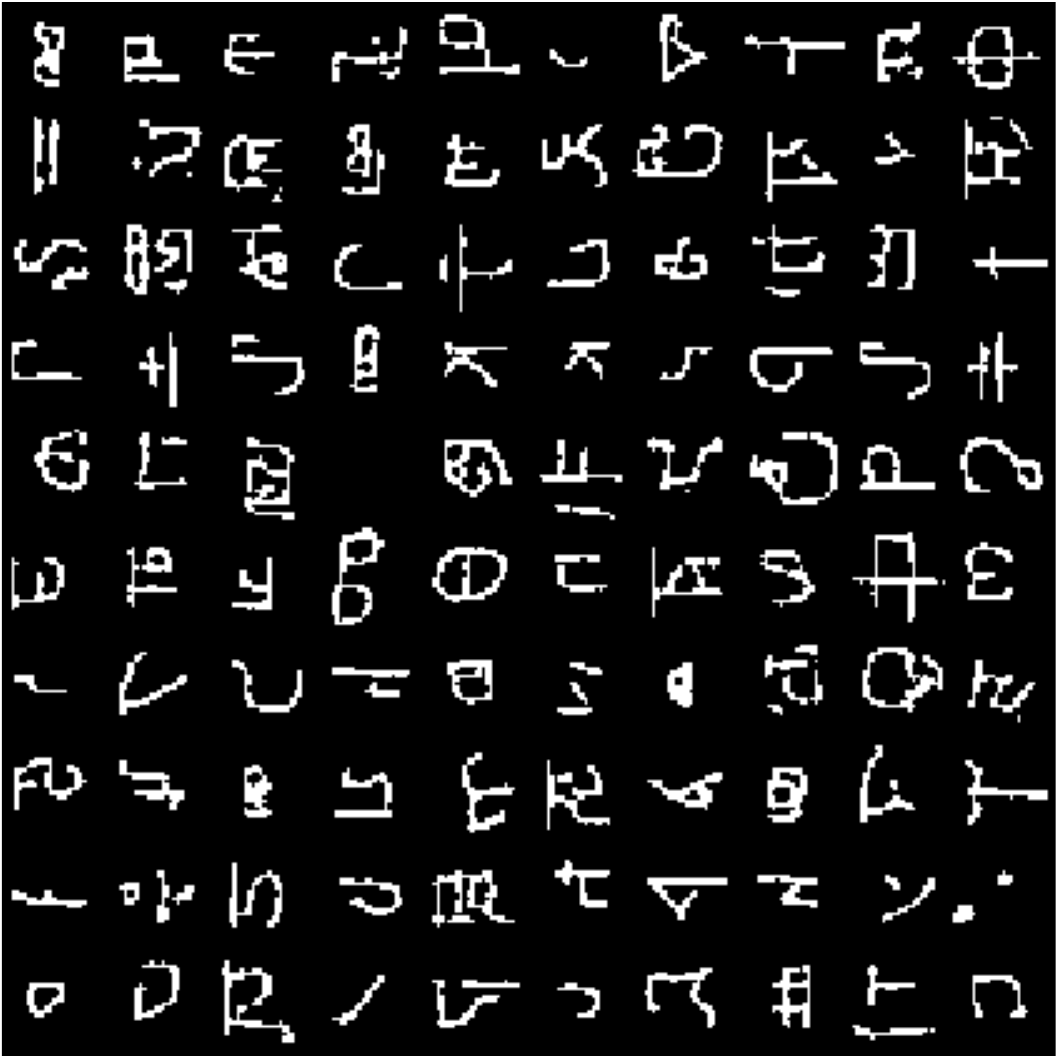}
    \caption{Reconstruction}
  \end{subfigure}
  \begin{subfigure}[t]{0.15\textwidth}
    \includegraphics[width=\textwidth]{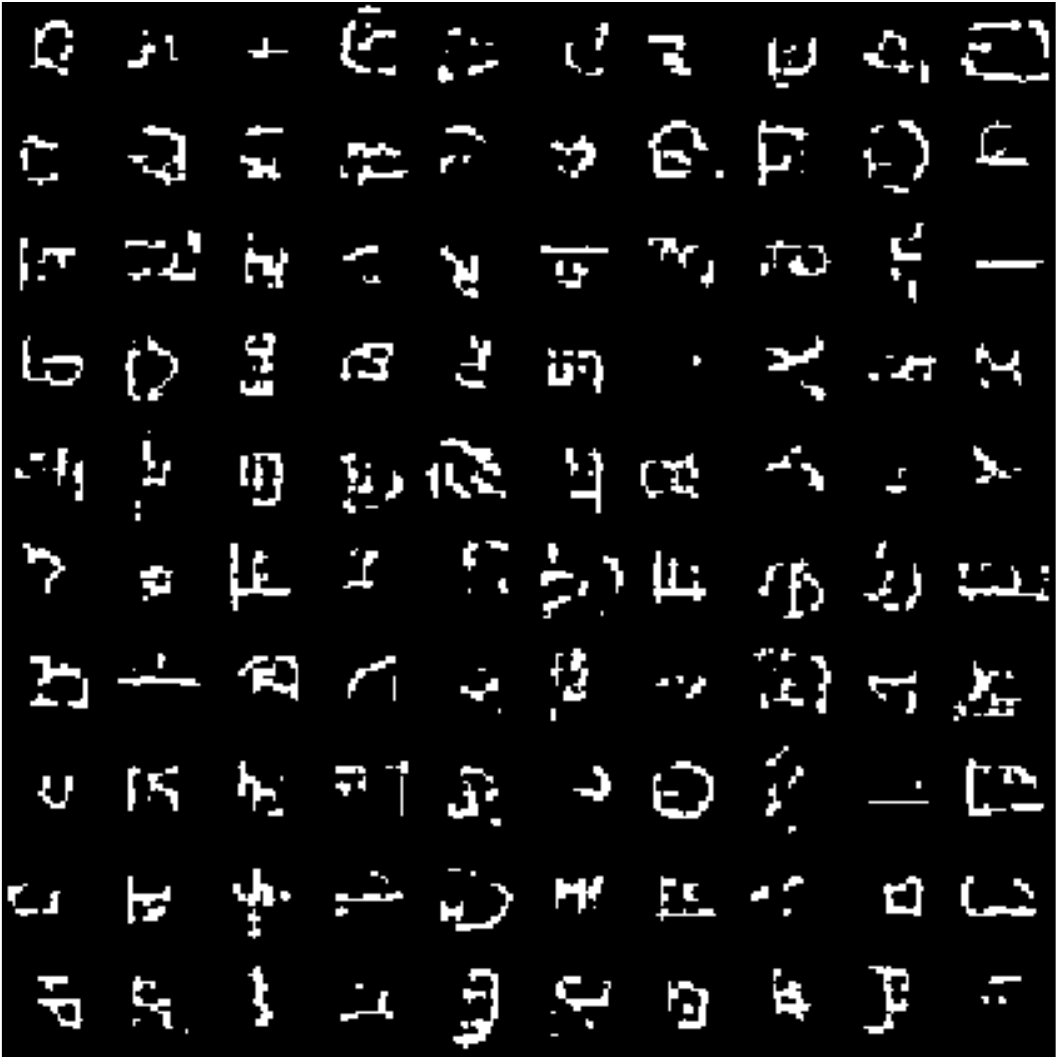}
    \caption{Random samples}
  \end{subfigure}
  \caption{Training data, its reconstruction and random
    samples. (Upper: MNIST, Lower: OMNIGLOT)}
  \label{fig:samples}
  \vspace{-0.15in}
\end{figure}

Generative samples can be obtained from trained model by feeding
$z\sim N(0,I)$ to the learned generative model $G_1$ (or $z_2\sim
N(0,I)$ to $G_2$). Since higher log-likelihood estimates are obtained
on $G_2$, Figure \ref{fig:samples} shows real samples from the
dataset, their reconstruction, and random data points sampled from
AVAE trained on $G_2$ for both MNIST and OMNIGLOT. We observe that the
reconstructions align well with the input data and that random samples
generated by the models are visually consistent with the training
data.

\section{Conclusions}
\label{sec:conclusion}

This paper presents AVAE, a new framework to enrich variational family
for variational inference, by incorporating auxiliary variables to the
inference model. It can be shown that the resulting inference model is
essentially learning a richer probabilistic mixture of simple
variational posteriors indexed by the auxiliary variables. We
emphasize that \textit{no density evaluations are required for the
  auxiliary variables}, hence neural networks can be used to construct
complex implicit distribution for the auxiliary variables. Empirical
evaluations of two variants of AVAE demonstrate the effectiveness of
incorporating auxiliary variables in variational inference for
generative modeling.

\bibliographystyle{named}
\bibliography{ref}

\end{document}